\documentclass{svproc}
\usepackage[T1]{fontenc}
\usepackage{url}

\begin{document}
\mainmatter              % start of a contribution
\title{Robots and Social Sustainability\thanks{This research was supported in part by the National Science Centre, Poland, under the OPUS call in the Weave programme under the project number K/NCN/000142.}}
\author{Bipin Indurkhya%\orcidID{0000-0002-3798-9209} 
\and Barbara Sienkiewicz%\orcidID{0009-0008-1977-5806}
}
\authorrunning{B. Indurkhya, B. Sienkiewicz}
\institute{Cognitive Science Department, Jagiellonian University, Krakow, Poland.
\email{bipin.indurkhya@uj.edu.pl; barbara.sienkiewicz@student.uj.edu.pl}}
\maketitle
\vspace{-3 mm}
\begin{abstract}
Sustainability is no longer a matter of choice but is invariably linked to the survival of the entire ecosystem of our planet Earth. As robotics technology is growing at an exponential rate, it is crucial to examine its implications for sustainability. Our focus is on social sustainability, specifically analyzing the role of robotics technology in this domain by identifying six distinct ways robots influence social sustainability.
\end{abstract}
\section{Introduction}
Sustainability is no longer an option but a dire necessity. One approach to achieving sustainability is to leverage existing technology \cite{bashir2022}. We focus here on how the technology of social robots relates to sustainability. We use the term social robot somewhat broadly to include chatbots and avatars. In recent years, there has been an increased interest in this area: a workshop on sustainability was organized as part of the IROS conference in 2021 \cite{IROS_2021,Mai2022}, and a session on sustainability was held at the European Robotics Forum in 2023 \cite{ERF_2023}. Nonetheless, sustainability has many facets, and the technology of social robots interacts with it in myriad ways. Here, we identify and analyze some of these interactions.

Sustainability, in general, refers to maintaining the capital, whether it be human capital, social capital, economic capital, or environmental capital \cite{Goodland_2002}. Our focus here is on social sustainability --- how the robotic technology can facilitate maintaining the social capital, which includes family relationships, friendships, group dynamics, and so on.
%As noted by \cite{kohl2020}, social sustainability is closely related to the ecological and the economic aspects.
We identify six ways in which social robots affect social sustainability: facilitating social inclusion, facilitating physical and mental health, provide companionship, act as mediators, educate about sustainability, and nudge towards sustainable behaviour, which are discussed below in turn.

\section {Robots facilitating social inclusion}

One major area where social robots are contributing to social sustainability is in facilitating social inclusion. Social robots are deployed for the purposes of communication, emotion expression and perception, social relationship maintenance, interpretation of natural cues, and the development of social competencies \cite{li2011towards,fong2003survey}. These skills are useful for people with neuro-developmental disorders such as autism \cite{pennisi2016autism} or ADHD \cite{cervantes2023social}. It has been shown that children with autism are more willing to talk with social robots than with other children \cite{scassellati2012robots}, which can be used to facilitate their social interaction skills.
Social robot can be a facilitator not only in one-on-one interaction, but also during group activities \cite{vanderborght2012using}. A robot Probo was used in the assisted therapy with the children with autism: the robot tells the stories through which it teaches how to react in social situations like `hello', `thank you'.

Incorporating people with disabilities into society is a fundamental step towards achieving equality and sustainability. It creates a sense of purpose and belonging for all individuals. Robotic avatars, as demonstrated by Barbareschi \cite{Barbareschi}, play a vital role in this process by not only promoting inclusiveness but also providing opportunities for the disabled to play a role (e.g. a waiter) that would not be possible due to their limitations. This not only enriches their lives but also contributes positively to the society as a whole.

One example of a social robot adding to the quality of life of visually disadvantaged people is provided by \cite{Kuribayashi}: a map-less navigation system {\it Pathfinder} is designed for blind people to reach destinations in unfamiliar buildings, which is implemented on a robot.

Another group that can benefit from interaction with social robots is children. Using humanoid robot during the rehabilitation of children with cerebral palsy increases their motivation and keeps them focused for longer \cite{kachmar2014humanoid,marti2018adapting}. Children under medical treatment at the hospital often feel stressed and need additional emotional support. Social robot can provide distraction during the medical procedure, and emotional or well-being support during the hospital stay when parents cannot be there \cite{moerman2019social}.
Social robots can also be used to increase awareness of hidden biases and discrimination. For example, in \cite{Romain}, a robot uses storytelling to teach children about gender equality.

In spite of all these approaches, we must be cautious in  evaluating their full impact, as it needs to be considered from the point of view of disabled people themselves \cite{shew2023}. Moreover, issues related to potential psychological harm from robot rejection must also be considered \cite{nash2018,spisak2023}.

\section {Robots for physical and mental health}

Another area where social robots have been effectively deployed to improve quality of life is in the care of the elderly. Cooper et al. \cite{Sara} point out the needs of our aging society, and how assistive robot can provide physical support and cognitive support. An interesting case in point is the Primo Puel robotic doll, which was introduced in Japan around $2004$. This doll had simple functional features --- it talks, giggles, and asks for cuddles --- and was designed as a companion for young single girls in the workforce. Perhaps surprisingly, these dolls were a big hit with the elderly population \cite{PrimoPuel}. They were found to be effective in monitoring the health and safety of the elderly, and to provide them companionship. Nonetheless, Sharkey and Sharkey \cite{Sharkey_2012} raise a number of ethical concerns related to such technologies, such as deception and infantilisation.

Another example is Paro the seal, which also has a simple design based on a baby harp seal. Covered in antibacterial fur, it can move its head and flippers a little bit. It can purr like a cat and its eyes can open and close. Its design has remained largely unchanged since its inception about twenty years ago. Paro the seal has been used effectively as a therapy aid for different user groups such as the elderly \cite{Birks_2016,bemelmans2015effectiveness},
people with dementia \cite{Kang_2020,Pu_2020,moyle2018care},
and cancer patients \cite{Eskander_2013}. However, 
even though Paro has been commercialized and used in care settings for more than a decade in multiple countries \cite{hung2019benefits}, to adequately assess its impact on quality of life, more studies are needed with larger sample sizes and a rigorous study design \cite{lihui2019}. Moreover, as with Primo Puel dolls, some ethical concerns have been raised about the use of this technology, such as self-deception \cite{calo_2011}.

Researchers have found that elderly users feel certain threats (the need for independence, the need for control, the fear of being replaced, and the need for authenticity) from robots \cite{deutsch2019home}, and face a number of psychological vulnerabilities \cite{liu2021can}.

%Any benefits that social robots may bring towards the quality of life for the elderly need to balanced by some potential negative effects, like causing anxiety \cite{deutsch2019home}, though this depends on cultural factors as an opposite effect was found with Chinese participants \cite{liu2021can}).

In recent years, many social robots and chatbots have been introduced as conversational agents that are especially designed for a therapeutic use. For example, the chatbot {\it Woebot} helps to deal with substance abuse \cite{Prochaska_2021}, and the chatbot {\it Replika} is designed to be an accepting, understanding, and non-judgmental companion \cite{xie2022attachment}.

However, some negative effects of these conversational agents have also been reported. For example, {\it Replika} has been blamed for inciting murder and suicide \cite{Possati_2022}. It has also been argued that chatbots can lead to addiction, thereby harming one's ability to form real-life intimate relationships \cite{xie2022attachment}. Some other ethical concerns of incorporating emotional response in social robots and chatbots are discussed in \cite{indurkhya2023}.

\section {Robots as companions}

For neurotypical children, social robots can facilitate their cognitive development, boost their creativity \cite{ali2019can,kory2019assessing}, and teach computational thinking \cite{yang2022towards}.

A more controversial issue is when social robots are treated as romantic partners. One such case is presented in Isabella Willinger’s 2019 documentary {\it Hi, A.I.}, where Chuck, a man from Texas is attempting to start a romantic relationship with a humanoid robot Harmony. Cindy Friedman \cite{friedman2023} raises some crucial ethical concerns about such use of technology from an African philosophical perspective. One obvious problem with such sexbots is that they promote addiction \cite{Boot_2018}. A number of other ethical issues related to this technology are being hotly debated \cite{Gonzalez_2021,Masterson_2022,saetra2022}.

Similarly, \cite{rasouli2022potential} discusses the potential use of social robots for coping with the social anxiety disorder, which is promising given the effectiveness of using social robots in interaction with children with ASD.
However, there is a risk of harmful side effects: for example, the companion social robot might become the only friend for the person with a high social anxiety, thereby causing further isolation.

Another way in which social robots are likely to make a significant impact on the quality of life in the near future is in doing household chores \cite{lehdonvirta2023}. However, as the study by Vaussard {\it et al.} \cite{vaussard2014} on the impact of robotic vacuum cleaners on home ecosystem has shown, one needs to look at the bigger picture by studying how such technology integrates with the user's space and perception to assess its overall impact.

%To summarise the discussion of this section, we see that there are many ways in which social robots can add to the quality of life a human user, however there are many potential ways in which such systems can be harmful, so extreme care is needed in deploying such systems, and their use requires constant monitoring.
 
%Social robots have been used in schools as a tutor of a second language \cite{kumar2020toy}; and as a peer or buddy to reduce stress levels in a real-life primary school environment \cite{smakman2022trustworthy}.

\section{Robots as mediators}

Social robots have also been found to be effective in mediating negotiations. For example, Druckman {\it et al.} \cite{druckman2021} compared the effectiveness of three different mediator platforms (teleoperated robotic, human, and computer screen-based) on negotiation outcomes, and found the robotic platform to be the most effective. It also generated more agreements outside of the pre-negotiation scenarios, and led to a more positive perception of the negotiation experience for the participants. 

In another recent study, Weisswange {\it et al.} \cite{weisswange2023}
have explored how social robots can be used for mediating complex social relationships in the real world. They have identified six different roles for robot mediators, and discussed mediation behaviours and measures to evaluate the quality of mediation interventions.

\section{Robots educating about sustainability}

Examining the intersection of robotic technologies with education and sustainable development is not new \cite{dias2005}. A more recent view is provided in \cite{park2016}. We briefly present a few example here.

Li {\it et al.} \cite{Li} address culturally-responsive computing (CRC) by introducing the concept of co-creation with a robot agent. In this approach, the robot agent itself is co-created through participatory design with adolescent girls and a social robot agent. It identifies themes about who has the power to make decisions, what decisions are made, and how to maintain social relationship.

A recent case study considers the problem of solving the waste problem at multi-day music festivals \cite{vella2023}. By using co-design workshops, the authors explored sociotechnical strategies to solve this problem together with the festival participants. As a result, they identified the challenges and opportunities for sustainable human-computer interaction on carving out a design response to a difficult problem by situating relations, meaning making, telling invisible stories, and finding leverage points.

In another example, Ferreira and Mi{\v{s}}kovic \cite{ferreira2021addressing} use robot as a tool to share magnificent underwater life with the goal that this will create a desire to preserve this environment. Also, social robots can help build good habits like hand hygiene \cite{sasidharan2021haksh,deshmukh2023}. Robots contribute to sustainable development not only through explicit teaching but also by effectively addressing and alleviating education challenges, such as the use of social robots in school anti-bullying programs \cite{sanoubari2021can}.

\section{Robots nudging towards sustainable behaviour}

Nudging \cite{Thaler_2008,Halpern_2015}, also called  Shikakeology \cite{Matsumura_2015}, means
changing the behavior of people by design.
A typical use case is to decide what you want people to choose, and to set that as the default option. This is because most people go with the default option. For example, if the default option is to be an organ donor (after your death) then there are many organ donors. Other examples include painting religious icons on a public wall to discourage men from urinating against it.

To promote socially responsible, sustainable behavior, nudge theory has been successfully applied in many areas in many countries around the world \cite{Sunstein_2019}. However, other studies have found that nudges do not always work, and sometimes end up nudging people in a direction opposite to the one that was intended \cite{Dimant_2020}. Engelen  \cite{Engelen_2019} found that some health-promoting nudges are not ethical. It has also been argued that nudging violates personal autonomy \cite{Schmidt_2020}.

There are several ways in which nudging can be used to promote sustainable behaviour \cite{gossen2022,lehner2016nudging,kasperbauer2017permissibility}.
One study specifically explored the effect of different types of agents in promoting normative behaviour \cite{tussyadiah2019nudged} and found that a robot by its presence itself can induce normative (sustainable) behavior. 

However, there are two major problems. One is that nudging, no matter how benevolent and how much based on good intentions, is essentially a kind of manipulation, and manipulation erodes trust \cite{Choo_2022}. It has been argued that nudging is incompatible with informed consent \cite{Simkulet_2019}. Another study \cite{Arcand_2007} focused on how people felt about opt-in format (by default your data is not shared, but you have to check a box to share) vs opt-out format (by default your data is shared, but you have to check a box to not share) in privacy statements. They found that participants who read the opt‐in format felt significantly more control and trust than participants who read the opt‐out format. Generally, people resent being manipulated, and if they sense it, they tend to go against the nudge, even at the risk of harming themselves.

Another issue is that any nudge creates a backdoor for commercial or political exploitation, as it depends on the intentions and transparency of the creators. Examples of unethical use of nudges include misleading wording, and order and presentation bias, to persuade the user for hasty purchases \cite{wendel2020designing}.

\section{Conclusions}

We have identified six aspects of social sustainability where robots are having a major impact. In many ways, robots are facilitating social interaction, especially in spanning across space and time limitations, and also overcoming cognitive limitations such as from Autism spectrum disorder (ASD). However, the technology for social robots is also creating alienation leading towards an emotionally dysfunctional society such as imagined in \cite{torras2018}.

\bibliographystyle{splncs04}
\bibliography{bibliography}
\end{document}